\documentclass[11pt,a4paper]{article}
\usepackage[hyperref]{ranlp2023}
\usepackage{times}
\usepackage{latexsym}

\usepackage{microtype}

\aclfinalcopy
\def\aclpaperid{58}

\usepackage{tcolorbox}
\usepackage{url}
\usepackage{booktabs}
\usepackage{multirow}
\usepackage{amsmath}
\usepackage{amssymb}
\usepackage{hyperref}

\title{Enhancing Transformer-Based Rerankers with Synthetic Data and LLM-Based Supervision}

\author{Dimitar Peshevski\textsuperscript{1, 2}\textsuperscript{ *}, Kiril Blazhevski\textsuperscript{1, 2}\textsuperscript{ *}, Martin Popovski\textsuperscript{1, 2}, \and Gjorgji Madjarov\textsuperscript{1, 2} \\ \\
        \textsuperscript{1}Faculty of Computer Science and Engineering, Ss. Cyril and Methodius University \\ Rugjer Boshkovikj 16, 1020, Skopje, North Macedonia \\ \\
        \textsuperscript{2}Machine Learning Department, G+D Netcetera \\ Partizanski Odredi 70b, 1020, Skopje, North Macedonia \\ \\
        \texttt{\normalsize dimitar.peshevski@\{finki.ukim.mk,netcetera.com\}},\\
        \texttt{\normalsize kiril.blazhevski@\{students.finki.ukim.mk,netcetera.com\}},\\
        \texttt{\normalsize martin.popovski@\{students.finki.ukim.mk,netcetera.com\}},\\
        \texttt{\normalsize gjorgji.madjarov@\{finki.ukim.mk,netcetera.com\}}}

\date{}

\begin{document}

\maketitle

\def\thefootnote{\textsuperscript{*}}\footnotetext{These authors contributed equally to this work.}\def\thefootnote{\arabic{footnote}}

\begin{abstract}
Effective document reranking is essential for improving search relevance across diverse applications. While Large Language Models (LLMs) excel at reranking due to their deep semantic understanding and reasoning, their high computational cost makes them impractical for many real-world deployments. Fine-tuning smaller, task-specific models is a more efficient alternative but typically depends on scarce, manually labeled data. To overcome this, we propose a novel pipeline that eliminates the need for human-labeled query-document pairs. Our method uses LLMs to generate synthetic queries from domain-specific corpora and employs an LLM-based classifier to label positive and hard-negative pairs. This synthetic dataset is then used to fine-tune a smaller transformer model with contrastive learning using Localized Contrastive Estimation (LCE) loss. Experiments on the MedQuAD dataset show that our approach significantly boosts in-domain performance and generalizes well to out-of-domain tasks. By using LLMs for data generation and supervision rather than inference, we reduce computational costs while maintaining strong reranking capabilities.
\end{abstract}

\section{Introduction}

Rerankers are essential in modern Information Retrieval (IR) systems, especially in Retrieval-Augmented Generation (RAG) pipelines, where initial retrieval often includes false positives. These irrelevant results can harm downstream tasks. Rerankers address this by refining search results—filtering out non-relevant documents and reordering based on relevance, thereby improving precision and system effectiveness.

Reranking methods range from traditional statistical models like BM25 to neural and transformer-based models. Transformer-based rerankers, such as those built on BERT, outperform others by capturing complex query-document relationships. Large Language Models (LLMs) can also be used for reranking due to their deep contextual understanding, but come with challenges like high latency and computational cost. In contrast, dedicated reranking models strike a better balance between efficiency and accuracy for real-time use.

However, rerankers require domain-specific fine-tuning to perform well. This typically needs labeled query-document pairs, which are often unavailable in specialized domains. To overcome this, we propose a query-less fine-tuning approach for cross-encoders using only a domain-specific text corpus. Our method uses an LLM to generate synthetic queries and identify relevant documents, creating synthetically labeled data for fine-tuning. This allows cross-encoders to approach LLM-level reranking performance without needing manually labeled data.

Our framework combines contrastive learning, knowledge distillation, and fine-tuning to adapt rerankers efficiently. By removing the need for curated relevance labels, it enables high-quality domain-specific reranking and improves retrieval precision where traditional rerankers fall short.

\section{Related Work}

As a task, reranking involves refining the initial list of retrieved documents by reordering them based on a model's relevance scores. The progress of deep learning has significantly advanced reranking approaches by enabling models to learn complex patterns in language. Recent work has demonstrated that transformer-based models, particularly the Bidirectional Encoder Representations from Transformers (BERT) model and its derivatives, offer state-of-the-art performance in reranking and other natural language processing tasks due to their strong contextual representation capabilities \cite{devlin2018bert}.

\subsection{Transformer-Based Models}

The introduction of transformer-based models has transformed the field of reranking. Models like BERT, RoBERTa \cite{roberta}, and T5 \cite{t5} significantly outperform traditional IR models, especially when fine-tuned for reranking tasks.

\citet{rerankingbert} first illustrate the potential of BERT for reranking in the context of passage retrieval, demonstrating substantial gains over prior methods on benchmarks like MS MARCO \cite{msmarco}.

Recent studies have also explored fine-tuning T5 for reranking, with \citet{nogueira-etal-2020-document} training it as a sequence-to-sequence model to generate relevance labels as "target words". The logits of these words are then interpreted as relevance probabilities, enabling ranking. Moreover, \citet{rankt5} adapt T5 with ranking-specific losses to directly compute ranking scores for query-document pairs. In particular, they explore pairwise and listwise ranking losses, achieving significant performance improvement across different datasets. Their results show that models fine-tuned with listwise ranking loss performed well in zero-shot scenarios on out-domain data, compared to models trained with classification losses.

\subsection{Fine-Tuning Techniques}

Fine-tuning techniques play a significant role in adapting pre-trained transformers for reranking tasks. \citet{moreira2024enhancing} present an ablation study analyzing reranking model performance based on model size, loss functions, and self-attention mechanisms.

The dominant approach to reranking involves task-specific fine-tuning, typically using pointwise ranking objectives to directly optimize relevance scoring for tasks like search, fact-checking, and question answering. This enhances the model’s ability to detect subtle relevance differences. Data augmentation—such as generating synthetic query-document pairs—further improves model robustness and generalization.

Contrastive learning in pair with knowledge distillation has the potential to make rerankers more efficient. While most such existing approaches focus on training or fine-tuning retriever models, these techniques can similarly be used for training or fine-tuning a ranking model. \citet{simcse} apply contrastive learning to maximize the distinction between relevant and irrelevant documents, resulting in more discriminative embeddings. Knowledge distillation, as shown in \citet{distilbert}, enables large models to be distilled into smaller, more efficient models without significant loss in accuracy, supporting deployment in resource-limited environments. This idea of distilling a larger and more capable model into a smaller and more specialized model can also be used in terms of ranking models.

\subsection{Contrastive Learning}\label{contrastive-learning}

Contrastive learning has gained attention as an effective technique for training retrieval and ranking models, particularly bi-encoders and cross-encoders, by helping the model learn more discriminative representations of relevant versus irrelevant documents. In training bi-encoders, the model achieves this by bringing relevant query-document pairs closer in the embedding space while distancing irrelevant ones. In training cross-encoders, the model scores each query-document pair independently by cross-referencing information between them, rather than encoding them separately.

A popular contrastive learning objective for reranking models is the InfoNCE (Information Noise-Contrastive Estimation) loss, introduced in \citet{oord2019infonce}. InfoNCE operates by maximizing the similarity between positive pairs (e.g., a query and its relevant document) while minimizing similarity to a set of negative examples (e.g., irrelevant documents). Using a categorical cross-entropy approach, InfoNCE identifies the correct positive sample among unrelated noise samples by probabilistically contrasting positives with negatives, encouraging the model to capture relevance distinctions. A key component of this process is negative sampling, where “hard negatives” or random negatives are used to focus the model on distinguishing relevance patterns, reducing overfitting to positive pairs alone.

In reranking tasks, using InfoNCE helps the cross-encoder learn relevance by focusing on both relevant and non-relevant document distinctions, which improves its ranking capability on diverse data. This approach allows reranking models to generalize effectively, even on challenging IR datasets, by embedding distinctions between relevant and irrelevant documents directly into the scoring function.

\subsection{Knowledge Distillation}

The Gecko approach demonstrates how knowledge distillation from large language models (LLMs) can effectively train a compact, high-performing embedding model \cite{lee2024gecko}. Gecko achieves competitive retrieval results through a two-step distillation process: first, generating synthetic data with an LLM, and then refining it by relabeling it with hard negatives. This distillation concept can also be applied to fine-tune ranking models, improving their efficiency and performance.

\subsection{Relevance Scoring}

To assess and rank retrieved passages effectively, various relevance scoring techniques are proposed nowadays.

\paragraph{Query Likelihood} \cite{sachan2023zeroshot} propose a ranking approach that improves passage retrieval in open-question answering by scoring relevance through a zero-shot question generation model. This approach uses a pre-trained language model to compute the likelihood of the input question given each retrieved passage, providing cross-attention without task-specific training and improving performance on unsupervised and supervised retrieval models alike. Additionally, recent advancements such as the ART approach introduced in \citet{sachan2023questionsneedtraindense}, enable effective unsupervised training of dense passage retrieval models by leveraging an innovative document-retrieval autoencoding scheme, which computes the probability of reconstructing the original question from retrieved evidence documents. This enables robust question and document encoding without labeled data, contributing to state-of-the-art results across multiple QA benchmarks.

\paragraph{Relevance Classification} \citet{zhuang2023beyond} propose an approach that improves zero-shot LLM-based ranking by incorporating fine-grained relevance labels in prompts, which enables more accurate differentiation between documents with varying levels of relevance to a query. This approach significantly improves ranking performance by reducing noise and bias, as shown on multiple BEIR datasets.

\paragraph{Cross-Attention Relevance} Meanwhile, \cite{izacard2022crossattn} present an approach that uses cross-attention scores from a reader model as relevance signals to train retrievers for tasks like question answering. By aggregating these attention scores across layers and heads, the approach generates synthetic labels for passage relevance without requiring annotated query-document pairs, achieving state-of-the-art results.

\subsection{LLMs as Rerankers}

LLMs have been adapted for document ranking using pointwise, pairwise, listwise, and setwise approaches, each balancing effectiveness and efficiency differently.

Pointwise ranking scores each passage independently, making it computationally efficient but less effective in capturing relative relevance. Pairwise ranking compares passages in pairs to improve ranking precision, but it requires multiple comparisons, increasing computational cost. Listwise ranking processes multiple passages in a single prompt, enhancing comparative relevance but consuming more tokens. Recent work, such as Listwise Reranker with a Large Language Model (LRL) \cite{ma2023listwise}, shows that zero-shot listwise ranking can outperform pointwise methods by directly generating reordered document lists without task-specific training.

Setwise ranking \cite{Zhuang_2024} presents a set of passages in a single prompt but focuses on selecting only the most relevant one, reducing LLM inferences while maintaining ranking effectiveness. This approach balances efficiency and effectiveness by combining the computational advantages of pointwise ranking with the comparative insights of pairwise and listwise methods. It achieves strong performance in zero-shot ranking tasks with lower token usage, making it a promising alternative for LLM-based document ranking.

\section{Methodology}

In this section, we describe the methodology for fine-tuning a cross-encoder on synthetic data for ranking tasks. We first explain our approach to generating queries from a corpus of passages using an LLM and detail the dataset preparation process. Then, we outline the training and evaluation procedures.

\subsection{Training Objective}

\textbf{Preliminaries} \quad Since we use a cross-encoder, we evaluate the relevance of a document \( d \) to a query \( q \) by passing both through an encoder-only model simultaneously and computing the relevance score as defined in Equation \(\ref{score-eq}\), where the symbol \( ; \) represents concatenation, \( cls \) extracts the vector representation of the [CLS] token and \( W_s \in \mathbb{R}^{d_{\text{model}} \times 1} \) is a weight matrix.

\begin{equation}\label{score-eq}
\begin{split}
    \text{score}(q,d) ={}& \text{cls}(\text{BERT}(\text{CLS};q_{1:n};\text{SEP}; \\
                         & \qquad d_{1:m}))W_s
\end{split}
\end{equation}

\textbf{Contrastive Loss} \quad To fine-tune the cross-encoder, we use the InfoNCE contrastive loss function, which is effective for optimizing relevance-based comparisons between positive and negative documents, as discussed in Section \ref{contrastive-learning}. This loss function enables the model to process both positive and negative documents for each query simultaneously, improving its ability to distinguish between them, particularly in the presence of challenging hard-negative examples.

The loss function encourages the model to assign higher similarity scores to relevant documents (\(d^+\)) than to non-relevant ones (\(d^-\)). Formally, treating \( \text{score}(q, d) \) as a deep similarity metric, the loss for a single query \( q \) is defined by Equation \ref{loss-eq}, where \( G_q = \{d_q^+\} \cup \{d_q^-\}_{i=1}^m \) represents the set of documents for the query, consisting of one positive document \( d_q^+ \) and \( m \) hard-negative documents \( d_q^- \).

\begin{equation}\label{loss-eq}
    \mathcal{L}_q \triangleq -\log \frac{e^{\text{score}(q, d_q^+)}}{\sum_{d \in G_q}{e^{\text{score}(q, d)}}}
\end{equation}

In our implementation, we use the \textit{Localized Contrastive Estimation (LCE)} approach proposed by Gao et al. \cite{gao2021rethink} to perform batch updates over a set of queries \( Q \). This method prioritizes hard negatives—documents that are both difficult and contextually relevant—over randomly selected negatives. These hard negatives are typically drawn from a pool of candidates ranked highly by a baseline retriever or another relevance scoring model. The batch-averaged loss is computed as defined in Equation \ref{batch-loss-eq}.

\begin{equation}\label{batch-loss-eq}
    \mathcal{L}_{LCE} \triangleq \frac{1}{|Q|} \sum_{q \in Q}{-\log \frac{e^{score(q, d_q^+)}}{\sum_{d \in G_q}{e^{score(q, d)}}}}    
\end{equation}

In the context of RAG systems, the retriever module constructs the candidate pool \( G_q \), which includes both relevant and hard-negative documents. Fine-tuning the ranking model with this loss improves its ability to distinguish between relevant and non-relevant documents among the top candidates retrieved in the initial stage.

\subsection{Synthetic Data Generation}

To effectively optimize the Localized Contrastive Estimation (LCE) loss, the training dataset must be structured as \( (q, d_q^+, \{d_q^-\}_{i=1}^m) \), where \( q \) is a query, \( d_q^+ \) is a positive document, and \( \{d_q^-\} \) represents a set of hard-negative documents.  

\begin{figure*}
  \centering
  \includegraphics[width=0.8\textwidth]{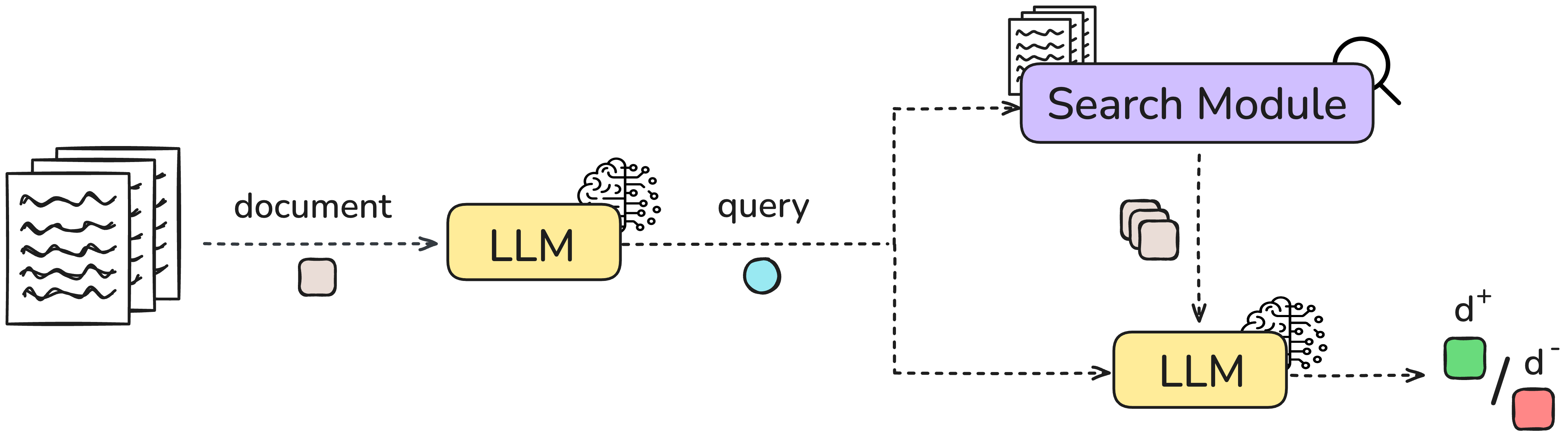}
\caption{Synthetic Dataset Generation Workflow: (i) Generate a synthetic query \( q \) from a given text corpus using a large language model (LLM). (ii) Retrieve the most relevant documents \( D_q \) for \( q \) from the corpus using a bi-encoder model. (iii) Evaluate each query-document pair in \( D_q \) using an LLM-based relevance scoring function \( f_{RC} \). (iv) Classify documents as relevant (positive) or irrelevant (negative) based on a predefined relevance threshold.}
  \label{fig:dataset-generation}
\end{figure*}

In scenarios where labeled query-document pairs are unavailable, such as in many RAG systems with pre-indexed documents, we address this challenge by generating the necessary data. Specifically, we: (1) generate synthetic user queries \( q \), and (2) mine positive and hard-negative documents for each synthetic query. We illustrate the steps involved in creating the synthetic dataset in Figure \ref{fig:dataset-generation}.

\subsubsection{Generating Synthetic Queries}

For a randomly selected document from the available corpus, denoted as \( d_{\text{seed}} \), and a carefully designed prompt \( \mathcal{P} \) tailored to the dataset's domain, we generate a synthetic query as: \( q = \text{LLM}(\mathcal{P}, d_{\text{seed}}) \). Since large language models (LLMs) perform well with few-shot prompting, we manually create several example queries from randomly selected documents in the corpus and incorporate them into \( \mathcal{P} \) to improve query generation quality.

\subsubsection{Mining Positive and Negative Documents}

To construct the set of one positive and \( m \) hard-negative documents for a synthetically generated query, we perform a search with \( q \) on the existing document corpus using a bi-encoder approach \cite{humeau2019poly}. This search produces a preliminary set \( D_q \) of top candidate documents for \( q \). We then apply a classification function to determine which of these candidates are positive (\( d^+ \)) and which are negative (\( d^- \)), forming the set \( G_q \) required for our loss function.  

Rather than relying on human annotators to obtain ground-truth labels for positive and negative documents from \( D_q \), we use a more powerful model as a teacher for the target ranking model. Specifically, we leverage an LLM-based relevance classification function to rerank the documents, assigning ground-truth labels in an automated manner.

\textbf{Relevance Classification} \quad Following the work of \citet{nogueira2020document} and \citet{zhuang2023beyond}, we use an LLM to estimate the probability of a document \( d \) being relevant by computing the log-likelihood of possible relevance labels given the query. Specifically, for a given query \( q \) and document \( d \), the LLM takes \( q \) and \( d \) as input and is prompted to respond with either "Yes" or "No" to indicate relevance. For each pair \( (q, d) \), where \( d \in D_q \), we compute the following score as defined in Equation \ref{relevance-eq}. This score corresponds to the probability assigned to the "Yes" token, computed by applying a softmax over the logits for the "Yes" and "No" tokens. The documents are then reranked based on this probability, with higher scores indicating greater relevance to the query.

\begin{equation}\label{relevance-eq}
    f_{RC}(q, d) \triangleq \frac{e^{LLM(Yes|\mathcal{P}, q, d)}}{e^{LLM(Yes|\mathcal{P}, q, d)} + e^{LLM(No|\mathcal{P}, q, d)}}
\end{equation}

\textbf{Mining Single Positive Passage} When generating synthetic queries using a large language model (LLM), it is reasonable to assume that the source document \( d_{\text{seed}} \), from which the query \( q \) is derived, is the most relevant document to \( q \). While this assumption generally holds, \citet{lee2024gecko} found that in approximately 15\% of cases, a different document is actually more relevant to \( q \) than \( d_{\text{seed}} \). This occurs because \( d_{\text{seed}} \) is sampled independently from the corpus without considering broader contextual relationships between documents.  

To address this issue, we redefine the positive document \( d_q^+ \) for \( q \) as in Equation \ref{positive-doc-eq}, rather than simply setting \( d_q^+ = d_{\text{seed}} \), which may lead to suboptimal training.

\begin{equation}\label{positive-doc-eq}
    d_q^+ = \underset{d \in D_q}{\operatorname{argmax}} \, f_{RC}(q, d)
\end{equation}

\textbf{Mining Multiple Negative Passages} Selecting negative documents from the candidate set \( D_q \) is more complex than identifying the positive document, as it involves selecting multiple documents rather than just one. We define the set of negative documents as those with a relevance classification score below a specified threshold as defined in Equation \ref{negative-docs-eq}.

\begin{equation}\label{negative-docs-eq}
    \{d_q^-\} = \{d \in D_q \mid f_{RC}(q,d) < t\}
\end{equation} 

The threshold \( t \) must be chosen carefully, as it can vary based on factors such as the LLM used, the prompt \( \mathcal{P} \), and dataset characteristics. In this study, we empirically determined that setting \( t = 0.5 \) produced effective results.

\subsubsection{Application in RAG Systems}

In retrieval-augmented generation (RAG) systems, the initial retriever accesses the knowledge base through an indexed vector store. Synthetic queries are generated from documents in this store, and the retriever retrieves a candidate set \( D_q \). The LLM-based relevance classification function then refines these candidates by identifying hard negatives and positives. This process enables the construction of a dataset suitable for fine-tuning the ranking model using a contrastive loss function.

\subsection{Evaluation Metrics}

To evaluate the ranking model before and after fine-tuning, we use standard ranking metrics that assess its ability to retrieve and rank relevant documents effectively \cite{croft2010search}. \( Precision@k \) measures the proportion of relevant documents among the top \( k \) retrieved results, indicating how well the model filters out irrelevant documents. Mean Average Precision (\( MAP@k \)) computes the mean of precision values at the ranks of relevant documents, rewarding models that rank relevant documents higher. Mean Reciprocal Rank (\( MRR@k \)) evaluates how early the first relevant document appears in the ranked list, with higher scores indicating better retrieval performance. Normalized Discounted Cumulative Gain (\( nDCG@k \)) quantifies ranking quality by considering both document relevance and position, applying a logarithmic discount.

\section{Experimental Setup and Evaluation}

\begin{table*}[h!]
    \centering
    \begin{tabular}{@{}cclcccccccccc@{}}
        \multicolumn{3}{c}{}&\multicolumn{10}{c}{\textbf{Dataset Length}} \\ 
        \cmidrule(lr){4-13}
        \rotatebox{90}{\textbf{Domain}} & \textbf{Metric} &  \textbf{Base}&\textbf{100} & \textbf{200} & \textbf{300} & \textbf{400} & \textbf{500} & \textbf{600} & \textbf{700} & \textbf{800} & \textbf{900} & \textbf{1000} \\ 
        \midrule
        \multirow{3}{*}{\rotatebox{90}{\textbf{In}}} 
            & MAP  &  0.911&0.915 & 0.919 & 0.933 & 0.938 & 0.938 & 0.939 & 0.942 & 0.944 & 0.940 & \textbf{0.946}\\ 
            & MRR  &  0.990&0.995 & 0.993 & 0.994 & 0.993 & 0.993 & \textbf{0.996}& 0.993 & 0.996 & 0.995 & 0.994 \\ 
            & NDCG &  0.905&0.913 & 0.922 & 0.945 & 0.949 & 0.945 & 0.950 & 0.950 & \textbf{0.952}& 0.946 & 0.951 \\ 
        \midrule
        \multirow{3}{*}{\rotatebox{90}{\textbf{Out}}} 
            & MAP  &  0.575&0.579 & \textbf{0.582}& 0.577 & 0.579 & 0.573 & 0.579 & 0.581 & 0.570 & 0.576 & 0.573 \\ 
            & MRR  &  0.575&0.579 & \textbf{0.582}& 0.577 & 0.579 & 0.573 & 0.579 & 0.581 & 0.570 & 0.576 & 0.573 \\ 
            & NDCG &  0.603&0.609 & \textbf{0.613}& 0.607 & 0.609 & 0.603 & 0.611 & 0.611 & 0.602 & 0.608 & 0.606 \\ 
        \bottomrule
    \end{tabular}
    \caption{Performance metrics ($@10$) across dataset lengths ranging from $100$ to $1000$, in increments of $100$, for in-domain (synthetic) and out-domain (subset of MS MARCO) data, evaluated after the first epoch of training.}
    \label{tab:performance}
\end{table*}

\begin{figure*}[h!]
  \centering
  \includegraphics[width=1\textwidth]{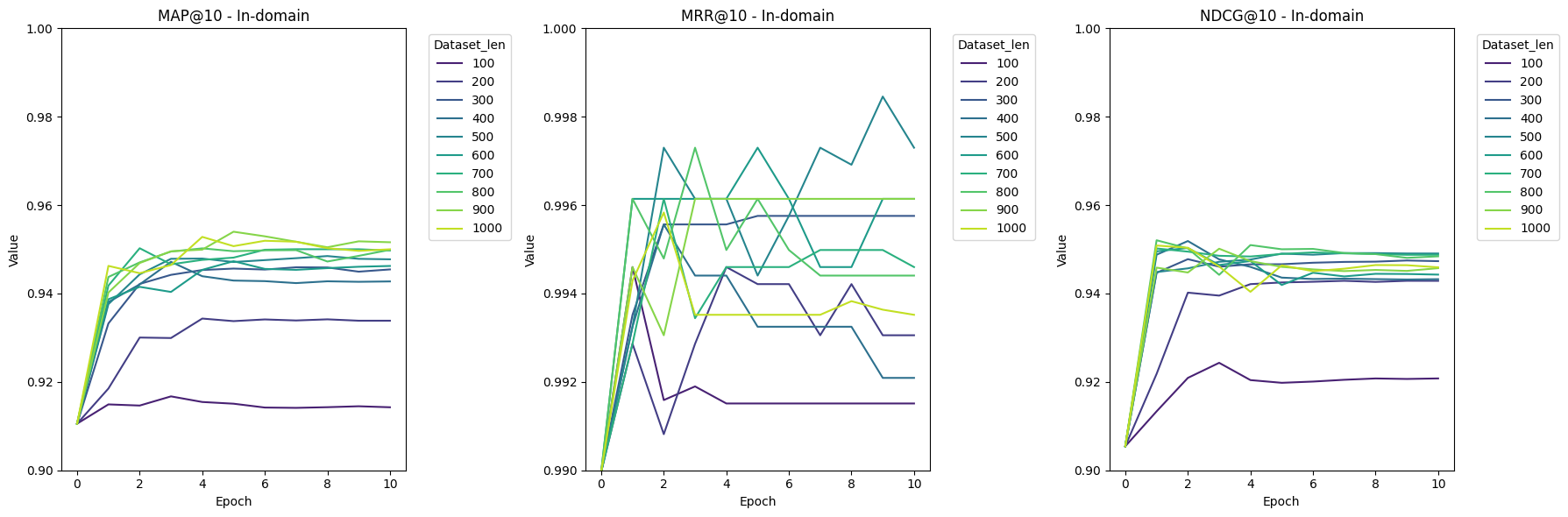}
\caption{Mean Average Precision (\(MAP@10\)), Mean Reciprocal Rank (\(MRR@10\)), and Normalized Discounted Cumulative Gain (\(NDCG@10\)) for the in-domain dataset, computed per epoch up to the 10th epoch. The training dataset sizes range from 100 to 1000 in increments of 100. These results demonstrate the effect of dataset size on model performance throughout the fine-tuning process.}
  \label{fig:in-domain-res}
\end{figure*}

\begin{figure*}[h!]
  \centering
  \includegraphics[width=1\textwidth]{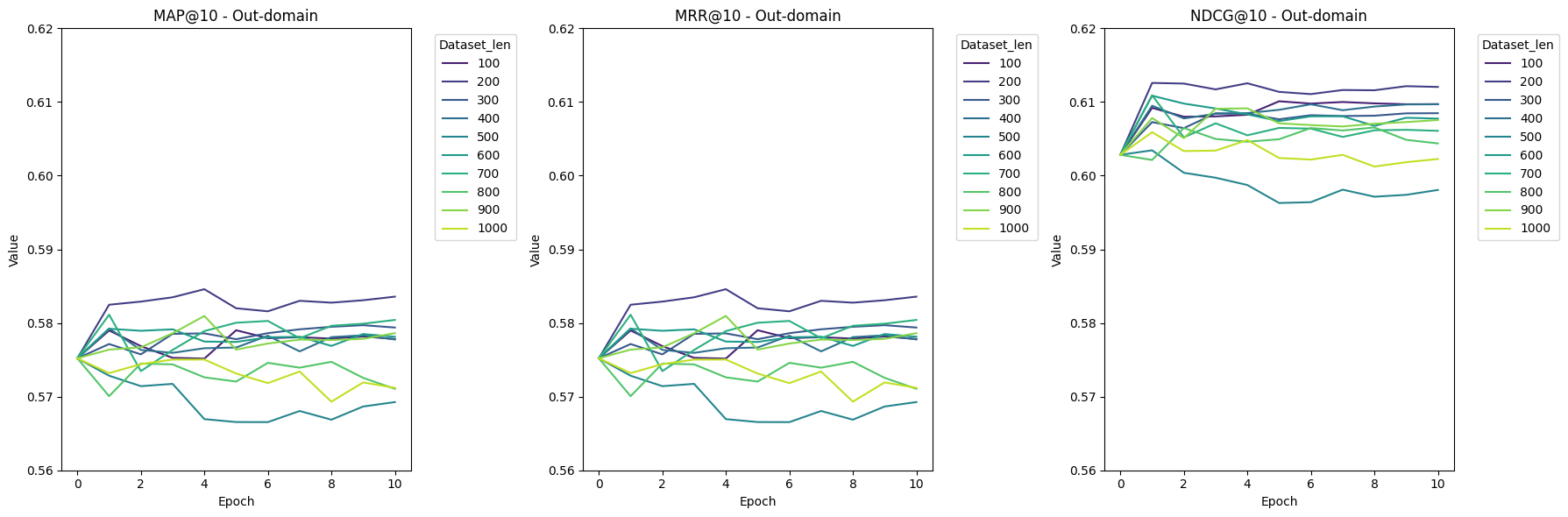}
\caption{Mean Average Precision (\(MAP@10\)), Mean Reciprocal Rank (\(MRR@10\)), and Normalized Discounted Cumulative Gain (\(NDCG@10\)) for the out-domain dataset, evaluated per epoch up to the 10th epoch. The training dataset sizes vary from 100 to 1000 in increments of 100, providing insights into the model's generalization performance during fine-tuning.}
  \label{fig:out-domain-res}
\end{figure*}

\begin{figure*}[h!]
  \centering
  \includegraphics[width=1\textwidth]{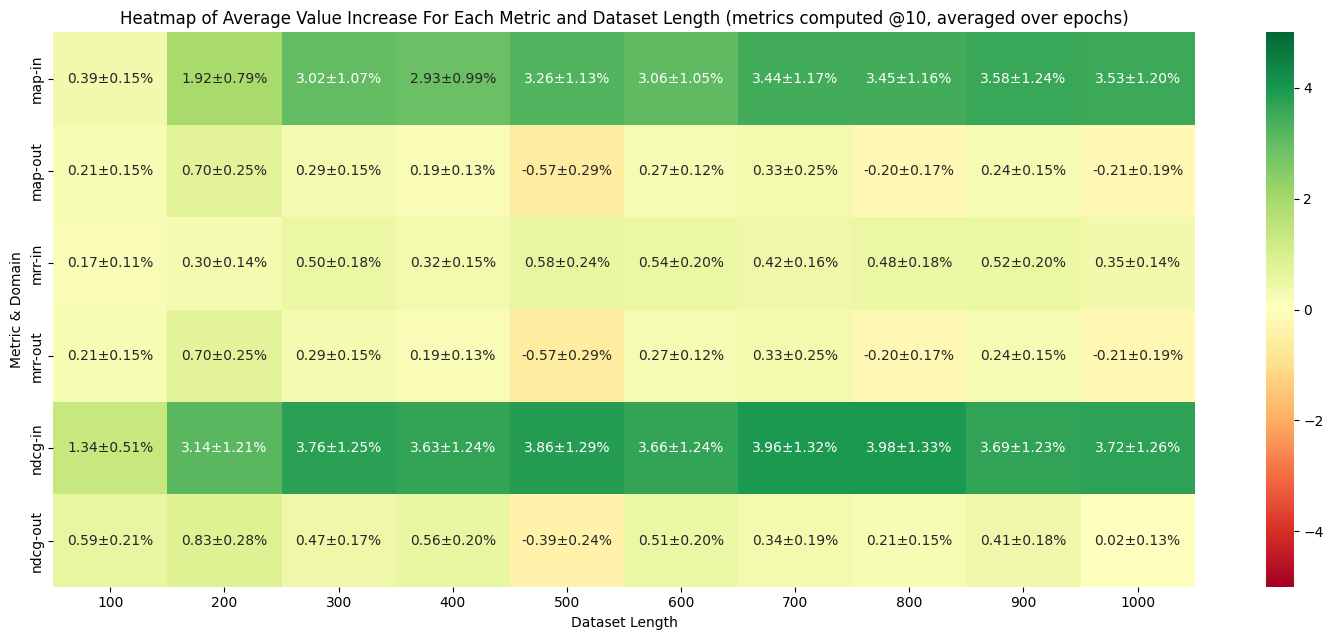}
\caption{Impact of training dataset size on performance metrics for both in-domain and out-of-domain datasets. The figure shows the average performance improvement (before vs. after training), computed across all epochs, with standard deviations representing variability.}
  \label{fig:heatmap}
\end{figure*}

\subsection{Generating Synthetic Queries}

To evaluate the proposed pipeline, we construct a synthetic dataset based on the MedQuAD dataset \cite{medquad} by generating queries (\( q \)) from its passages. MedQuAD consists of 47457 question-answer pairs extracted from 12 reputable NIH-affiliated websites, covering 37 question types across key medical topics such as treatments, diagnoses, side effects, diseases, medications, and medical tests.

To simulate a RAG setting—where only an indexed collection of documents is available—we treat the answers as standalone passages. Additionally, to accommodate the 512-token input limit of our target reranker, passages exceeding this length are excluded, and a random subset of 1000 answers is selected for synthetic query construction.

We use few-shot learning to guide the query generation process, ensuring the queries aligned with the domain of the passages. The few-shot examples used in the prompt are selected from the same domain, enabling the model to produce contextually relevant queries.

\subsection{Mining Positive and Negative Passages}

To identify positive and negative passages for each synthetically generated query \( q \), we use a bi-encoder model, specifically the \href{https://huggingface.co/intfloat/multilingual-e5-large}{\texttt{intfloat/multilingual-e5-large}} transformer. Both \( q \) and the passages are embedded using mean pooling to generate dense vector representations. We then retrieve the top 30 candidate passages from MedQuAD for each query using cosine similarity, forming the candidate set \( D_q \).  

We then perform relevance classification for each query-passage pair \( (q, d) \), where \( d \in D_q \), using a large language model, specifically a 4-bit GPTQ quantized version of \href{https://huggingface.co/hugging-quants/Meta-Llama-3.1-70B-Instruct-GPTQ-INT4}{Llama-3.1} with 70 billion parameters. Each passage is classified as positive or negative based on its relevance score, with a threshold of 0.5 used for classification.

\subsection{Dataset Preparation}

After constructing the triplets \( (q, \{d_q^+\}, \{d_q^-\}) \), we divide the dataset into training and test sets, with the test set containing 500 samples. To assess whether the dataset size is sufficient for fine-tuning, we conduct experiments using different subsets of the training data. The results of these experiments are presented in the following sections.  

The synthetic dataset represents in-domain data, aligning with the target domain where improved reranking performance is needed. To evaluate the model's generalizability and mitigate potential performance degradation on general knowledge tasks, a phenomenon known as \textit{catastrophic forgetting}, we include an out-domain dataset in our evaluation. This dataset consists of a subset of the MS MARCO dataset, and we monitor performance metrics on this dataset throughout the experiments.

\subsection{Fine-Tuning the Model}

To assess the effectiveness of our approach, we fine-tune the \href{https://huggingface.co/BAAI/bge-reranker-v2-m3}{\texttt{BAAI/bge-reranker-v2-m3}} model on different subsets of the synthetic dataset. The training subsets range from 100 to 1000 entries, with each subset being a strict superset of the previous one. This setup allows for a progressive evaluation of how dataset size impacts performance. The evaluation set remains the same across all experiments to ensure result comparability.  

For training, we use a batch size of 2 with gradient accumulation over 2 steps, yielding an effective batch size of 4. Each query \( q \) is paired with one positive and four negative passages during training. In evaluation, we increase the number of positive and negative passages to better reflect real-world scenarios while ensuring that the total number of passages per query do not exceed 30 to maintain computational efficiency.

\subsection{Results and Analysis}

We present the results of our experiments in Figure \ref{fig:in-domain-res} and Figure \ref{fig:out-domain-res}, showing that increasing the training dataset size improves model performance on the in-domain dataset. However, these improvements tend to plateau as the dataset size grows.  

The highest \( nDCG@10 \) score of 0.952 is achieved using a dataset of 800 entries after the first epoch, marking a good improvement from the initial score of 0.905. The best \( MAP@10 \) score is observed with a dataset of 900 entries, while the highest \( MRR@10 \) score is achieved with 500 entries.
The performance metrics across dataset lengths in increments of 100 elements, for both in-domain and out-domain data evaluated after the first epoch of training, are provided in table \ref{tab:performance}. The selection of the first epoch for reporting the evaluation metrics was informed by knee plot analysis, which indicated that for most dataset lengths, the most substantial performance gains occurred early in the training process.

Fine-tuning on the in-domain dataset does not result in any significant degradation in the model's performance on the out-domain dataset. In some cases, performance metrics on the out-domain dataset showed slight improvements, though these changes are not substantial (as shown in Table \ref{tab:performance}).  

To provide a comprehensive overview, in Figure \ref{fig:heatmap}, we illustrate the changes in performance metrics for both the in-domain and out-domain datasets as a function of training dataset size. We present the results averaged across all epochs, along with the standard deviations to indicate variability. This visualization highlights the effectiveness of fine-tuning the model on in-domain data while maintaining the same generalization to out-domain tasks.

\section{Conclusion and Future Work}

In this work, we propose a novel method for fine-tuning transformer-based reranking models without relying on manually labeled query-document pairs. Instead, we generate synthetic training data using large language models (LLMs), making the approach especially suitable for domain-specific applications lacking labeled data.

Our method uses contrastive learning with Localized Contrastive Estimation (LCE) loss to train the model to distinguish relevant from non-relevant documents. LLMs generate queries and evaluate relevance, automating the creation of training triplets. We further enhance learning by incorporating hard negatives, identified via bi-encoder retrieval and refined using LLM-based scoring.

This pipeline produces diverse, domain-specific data that boosts reranking performance in Retrieval-Augmented Generation (RAG) and other IR tasks. Experiments show significant improvements in retrieval quality.

Future directions include incorporating reinforcement learning to optimize synthetic data generation and extending the method to multilingual settings. We also explore integrating knowledge graphs to guide query generation, using structured domain knowledge to produce more relevant and less hallucinated queries.

By eliminating the need for manual labels, our method makes domain-specific reranking more scalable and effective, advancing ranking optimization in modern IR systems.

\bibliographystyle{acl_natbib}
\bibliography{references}

\end{document}